\pdfoutput=1
\documentclass[runningheads]{llncs}

\usepackage{graphicx}

\usepackage{times}
\usepackage{latexsym}

\usepackage[T1]{fontenc}
\usepackage[utf8]{inputenc}
\usepackage{microtype}
\usepackage{inconsolata}

\usepackage{graphicx}
\usepackage[utf8]{inputenc} %
\usepackage[T1]{fontenc}    %
\usepackage{hyperref}       %
\usepackage{url}            %
\usepackage{booktabs}       %
\usepackage{amsfonts}       %
\usepackage{nicefrac}       %
\usepackage{microtype}      %
\usepackage{xcolor}         %
\usepackage{caption}
\usepackage{subcaption}
\usepackage{cleveref}

\usepackage{float}
\usepackage{microtype}
\usepackage{graphicx}
\usepackage{arydshln}
\usepackage{longtable}
\usepackage{rotating}
\usepackage{tablefootnote}
\usepackage{multirow}
\usepackage{multicol}
\usepackage{booktabs}
\usepackage{makecell}
\usepackage{arydshln}
\usepackage{xspace}
\usepackage{color,soul}
\usepackage{tabularx}
\usepackage{inconsolata}
\usepackage{amssymb}
\usepackage{calc}
\usepackage{enumitem}

\usepackage{amssymb}
\usepackage{graphicx}
\usepackage{calc}

\usepackage{caption}
\usepackage{subcaption}
\usepackage{cleveref}

\newcommand{\bsm}{Seller Messaging\xspace}

\newcommand{\heuristicbaseline}{\mbox{\textsc{HB}}\xspace}
\newcommand{\approach}{\mbox{\textsc{M2Q}}\xspace}
\newcommand{\approachname}{\mbox{\textsc{Message-to-Question}}\xspace}
\newcommand{\hybridapproach}{\mbox{\textsc{\textsc{M2Q-Hybrid}}}\xspace}
\newcommand{\flanxxl}{\mbox{\textsc{FT5}}\xspace}

\newcommand{\flanxxlname}{{\textsc{flan-t5-xxl}}\xspace}

\newcommand{\vicuna}{\mbox{\textsc{Vicuna}}\xspace}

\newcommand{\vicunacmd}[2][1=]{{\textsc{Vicuna-#2}}\xspace}
\newcommand{\flancmd}[2][1=]{{\textsc{FT5-#2}}\xspace}
\newcommand{\approachcmd}[2][1=]{{\textsc{M2Q-#2}}\xspace}
\newcommand{\hybridapproachcmd}[2][1=]{\mbox{\textsc{\textsc{M2Q-Hybrid-#2}}}\xspace}

\newcommand{\control}{\mbox{\textsc{C}}\xspace}
\newcommand{\treatment}{\mbox{\textsc{T}}\xspace}
\newcommand{\treatmentrel}{\mbox{\textsc{T$_{pos}$}}\xspace}

\title{Instant Answering in E-Commerce Buyer-Seller Messaging using Message-to-Question Reformulation}

\titlerunning{Instant Answering in E-Commerce Buyer-Seller Messaging}

\footnotetext{These authors contributed equally to this work.}

\author{Besnik Fetahu$^\dagger$ \and
Tejas Mehta$^\dagger$ \and
Qun Song \and
Nikhita Vedula \and
Oleg Rokhlenko \and
Shervin Malmasi}
\authorrunning{Fetahu et al.}

\institute{Amazon.com, Inc. Seattle, WA, USA \\
\email{\{besnikf,mehtejas,qunsong,veduln,olegro,malmasi\}@amazon.com} 
}

\begin{document}

\maketitle

\begin{abstract}

E-commerce customers frequently seek detailed product information for purchase decisions, commonly contacting sellers directly with extended queries. This manual response requirement imposes additional costs and disrupts buyer's shopping experience with response time fluctuations ranging from hours to days.
We seek to automate buyer inquiries to sellers in a leading e-commerce store using a domain-specific federated Question Answering (QA) system. The main challenge is adapting current QA systems, designed for single questions, to address detailed customer queries.
We address this with a low-latency, sequence-to-sequence approach, \approachname (\approach). It reformulates buyer messages into \emph{succinct} questions by identifying and extracting the most salient information from a  message. 
Evaluation against baselines shows that \approach yields relative increases of 757\%  in question understanding, and 1,746\% in answering rate from the federated QA system.
Live deployment shows that automatic answering saves sellers from manually responding to millions of messages per year, and also accelerates customer purchase decisions by eliminating the need for buyers to wait for a reply.

\end{abstract}

\section{Introduction}
\label{sec:intro}

The rapid growth of e-commerce, with hundreds of millions of global users, hinges on customer satisfaction tied to easy, instant access to comprehensive product details. However, the available product descriptions, customer reviews, and QAs on e-commerce sites may not provide enough information for users to make informed purchasing decisions.

E-commerce platforms use a \textit{\bsm} feature for facilitating communication between buyers and sellers through an online messaging system about product details, shipping, or post-purchase concerns. However, large platforms which handle over 100K messages daily experience high wait times ranging from \emph{1-2 hours} to \emph{several days} due to the time and monetary costs of manual responses, thus delaying purchase decisions.

\begin{table*}[!ht]
    \centering
    \resizebox{1.0\linewidth}{!}{
    \begin{tabular}{p{6cm}@{\hskip .5cm}p{5cm}@{\hskip .5cm}p{5cm}}
        Buyer's Message to Seller & Ground Truth & Reformulated Message\\
        \toprule

         I'm trying to look into Forids trash bags, and have found that the website and facebook on the box dont exist :( \textcolor{red}{\textbf{\emph{I'm curious about their sourcing and material is used}}} !! & what is the sourcing material of this product? & \emph{what is the sourcing material used for \textcolor{black}{this product}}?\\
         
         \midrule

        Hello! I’m curious whether \textcolor{red}{\textbf{\emph{this Re:Zero REM figure you’re selling is from the authentic Taito Coreful brand?}}} Thank you. &  is this product from the taito coreful brand? & \emph{is this product from the authentic taito coreful brand?}\\
        
        \midrule

        My family surname is not a known surname. Are you  \textcolor{red}{\textbf{\emph{able to create a family crest with all the emblems and mottos?}}} Looking forward to hear from you. &  can this product be customized with all the emblems and mottos? & \emph{can this product be created with all the emblems and mottos}?\\
        
        \bottomrule
    \end{tabular}}
    \caption{Examples of buyer messages reformulated by \approach with the buyer's main intent in red.}
    \label{tab:examples}
\end{table*}

We address these gaps by leveraging a federated, multiple backend QA system for instantly answering customer questions using knowledge sources such as reviews, product catalogs, manuals, and community QA.
However, \emph{style mismatch} is a key challenge: customers write lengthy email-style messages, while traditional QA systems operate on simple direct questions.
We tackle this via an end-to-end approach, \approachname (\approach), using sequence-to-sequence models
to reformulate lengthy buyer messages
into short standalone questions.
\approach is optimized to distill relevant details from buyer's messages into \emph{answerable} questions and use a state-of-the-art QA model to generate precise answers. This allows the \bsm feature to provide instant QA by leveraging existing resources without incurring additional time or cost overheads (see Table~\ref{tab:examples}). During message reformulation, \approach considers the buyer's primary needs, combines multiple needs if necessary into a concise request for the QA system, and ensures customer privacy by omitting personal information.

\section{Background}\label{sec:background}

\textbf{Buyer-Seller Interactions \& QA:}
Prior research has highlighted the financial and commercial importance of studying interactions between buyers and sellers on e-commerce stores, as well as answering buyer inquiries in a fast and effective manner~\cite{masterov2015canary,kumar2019question,ahearne2022future}. 
We propose to improve buyers' access to instantaneous answers, and reduce product sellers' burden and expenses on e-commerce stores by integrating product question answering (QA) systems~\cite{gao2021meaningful,deng2023product} into the \bsm feature. 
Prior work on automatically answering customer queries within customer service applications, focuses on retrieving answers from a knowledge base~\cite{li2018question,cui2017superagent,liao2021practical,samarakoon2011automated} as well as generating answers~\cite{chen2020jddc,peng2023check}.

\textbf{Text Reformulation:}
Buyer messages can be verbose with long descriptions or irrelevant personal details (see Table~\ref{tab:examples}).
This distracts QA models and makes it difficult for them to provide accurate answers \cite{lyu2021zero,cao2022tasa,shi2023large}. 
\approach ensures that buyer message reformulations can \emph{adapt} to and maximize the understanding and answering rate of existing QA systems. 
Several Large Language Models (LLMs) have been fine-tuned to summarize and extract salient information from dialogues and email threads~\cite{feng2021survey,rennard2022abstractive}; identify the appropriate context to be input to a model to answer questions effectively~\cite{mcdonald2022detect,mao2020generation}; reformulate questions for easy answering~\cite{ferguson2022investigating,DBLP:conf/acl/FaustiniCFRM23,vakulenko2021question}; as well as to select the most relevant conversation history as context in case of conversational QA~\cite{zaib2022conversational,do2022cohs}. 

\section{\approach: \approachname}
\label{sec:approach}

\begin{figure*}[ht!]
    \centering
    \includegraphics[width=1\linewidth]{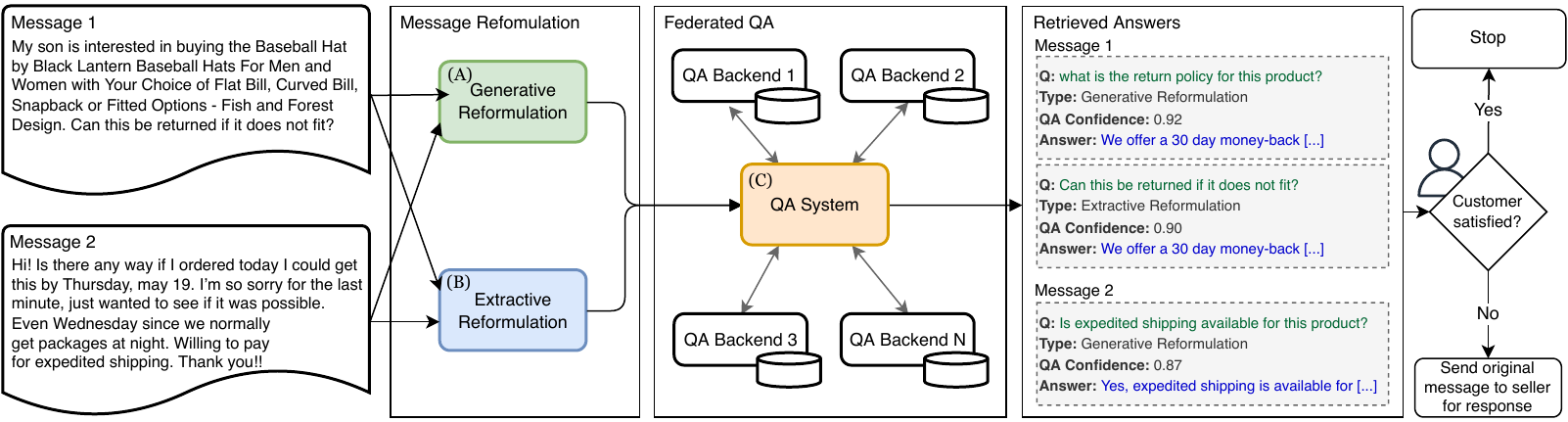} 
    \caption{\approach approach overview. For each buyer message that may contain multiple questions/intents, \approach reformulates them (using (B) and (C)) into a \emph{succinct question} with the most salient intent (or conjunction of intents). These are sent to a federated QA system (C) for instant answering.
    If customers are unsatisfied with the response or receive none, the original message is forwarded to the seller for manual reply. }
    \label{fig:rise_overview}
\end{figure*}

\approach reformulate buyer's messages into succinct questions that are instantly answered using a federated QA system.
If the question cannot be answered, or buyers are dissatisfied with the automatic response, they can forward their message to the seller for a manual response.
\Cref{fig:rise_overview} shows an overview of the approach. 
Messages sent to sellers are reformulated into questions, which is then used to retrieve an instant answer from the QA system.
Next, we describe the components from the figure.

\paragraph{\textbf{Generative Reformulation (A):}}
The main objective of this step is to convert buyers' lengthy messages into a shorter, concise form that correctly captures their primary intent, filtering out any irrelevant details or personal information. 
\Cref{tab:examples} shows examples of such reformulations. 
For the generative reformulation, we fine-tune different sequence-to-sequence models to perform the rewriting automatically, using a parallel dataset of buyer messages and their human generated reformulations (c.f. \S\ref{sec:data}).

\paragraph{\textbf{Extractive Reformulation (B):}}
A simpler reformulation method involves splitting the buyer message into sentences, and running them through a question classifier provided by the QA system. The sentence with the highest confidence is chosen to represent the entire message. This method works well for simple messages containing a direct question, but fails on more complex messages, and may not identify any questions at all.

\paragraph{\textbf{Question Answering System (C):}}
The reformulated messages from the previous steps are input to a federated QA system, which passes them to numerous in-house answer retrieval systems. This system returns answers and a \emph{QA confidence score}.
The QA system is a black box in this work; we do not modify or tune it.
We show in \S\ref{sec:evaluation} that shorter messages reformulated by \approach maximize the performance of downstream QA systems, leading to an increase in the automatic question answering rates.

\section{Dataset Construction}
\label{sec:data}

We created a dataset of $\sim$6k pairs of buyer messages and their target reformulations, divided into 5k/600/450 for train/dev/test sets.
Most messages are between 25-75 words.

\textbf{Annotation Guidelines.}
We engaged in-house annotators to transform messages into answerable questions based on certain guidelines, ensuring the quality of the parallel dataset\footnote{Our human annotators are expert in-house annotators that provide their relevance judgements based on a pre-determined annotation protocol, which was designed specifically for this task.}. First, they checked for English language and then identified the number of intents in each message. With multi-intent messages, they focused on the primary intent considering the context, while the remaining were most likely follow-up questions. The message was transformed into a precise question that pinpointed the buyer's main intent and information need. Table~\ref{tab:examples} shows example annotations.

\textbf{Annotation Quality Control.}
A secondary quality check is performed on all annotations by an expert annotator due to the subjective nature of summarizing buyer messages and potential for misinterpretation. The absolute agreement rate between expert and in-house annotators is 76\%. Discrepancies are corrected by the expert annotator.
\section{Offline Experiments and Results}
\label{sec:setup}

\subsection{Baselines and Our Approach}
\label{subsec:baselines}

We compare several reformulation methods\footnote{Due to privacy regulations, we cannot use external API-based LLMs like ChatGPT.} against the same QA system.

\paragraph{\textbf{Extractive Reformulation:}}
The sentence with the highest question classifier confidence score from the buyer message is selected. This works best with short messages that contain a clear, single question.

\paragraph{\textbf{Flan-T5-XXL:}} We assess \textsc{Flan-T5}'s reformulations  in zero- and few-shot (3 exemplars) settings.

\paragraph{\textbf{{Vicuna-13B}:}} We assess the Vicuna LLM \cite{vicuna2023}, which reportedly obtains 90\% of GPT-4's performance, in the same zero-shot and few-shot settings as \textsc{Flan-T5}.

\paragraph{\textbf{Approach:}}
We evaluate our \approach approach with two underlying base models. \approachcmd{T5} uses the smaller 220M parameter \texttt{T5-base} model \cite{raffel2020exploring}, while \approachcmd{FT5} uses \flanxxlname \cite{DBLP:journals/corr/abs-2210-11416} with 11B parameters. We consider two configurations:

\begin{itemize}
    \item \textbf{\approach:} The buyer's original message is converted into a standalone question via generative reformulation only, and sent to the QA system.
    
    \item \textbf{\hybridapproach}:
    We combine generative and extractive reformulation methods. The extractive approach responds when it obtains an answer, else generative reformulation is employed. \hybridapproach thus allows direct responses to shorter messages, decreasing error probability from reformulations.
    
\end{itemize}

\subsection{Evaluation Strategies}
\label{subsec:eval_strategies}

We measure reformulation quality in several ways:

\begin{itemize}
\item \textbf{Text Generation Performance}:
We use BLEU and ROUGE to measure how closely the revised messages align with their originals.

\item \textbf{Reformulation Accuracy:}
Human annotators evaluate reformulated messages based on binary score relevance, verifying the buyer's true intent.

\item \textbf{Understand Rate:} Measures the QA system's capability to understand buyer messages or the corresponding reformulations.

\item \textbf{Answer Rate:} Measures the answerability of buyer messages by our QA system.

\item \textbf{Answer Relevance:} 
We manually rate each message or its revision as either ``\emph{helpful}'' or ``\emph{unhelpful}'' based on the answer's correctness or if it's unanswered.

\end{itemize}
\subsection{Results}
\label{sec:evaluation}

\begin{table*}[ht!]
    \centering
    \resizebox{1.0\textwidth}{!}{
    \begin{tabular}{p{4cm} r | r r r r r r r | r r r}
         \toprule
         \multicolumn{1}{c}{} & \multicolumn{8}{c}{Generation Performance} & \multicolumn{3}{c}{QA Performance} \\
         \midrule\midrule
   & \makecell{Reformulation\\ Accuracy} & BLEU1 & BLEU2 & BLEU3 & BLEU4 & ROUGE1 & ROUGE2 & \multicolumn{1}{c}{ROUGEL} & \multicolumn{1}{|c}{\makecell{Understand\\Rate}} & \makecell{Answer\\ Rate}  & \makecell{Answer\\ Relevance} \\
 \midrule
Extractive Baseline  & 41\% & 0.143 & 0.060 & 0.036 & 0.025 & 0.228 & 0.099 & 0.214 & - & - & - \\[1ex]

\vicunacmd{Zero-shot} & - & 0.139 &  0.058 &  0.037 &  0.027 &  0.330 &  0.121 &  0.292 & +235\% & +827\% & +320\%\\
\vicunacmd{Few-shot}  & 31\% & 0.335 &  0.206 &  0.145 &  0.098 &  0.504 &  0.280 &  0.484 & +578\% & +1,182\% &  +1,280\%\\[1ex]

\flancmd{Zero-shot}  & - & 0.356 &  0.182 &  0.136 &  0.108 &  0.489 &  0.231 &  0.450 & +616\% & +1,246\% & +1,240\%\\
\flancmd{Few-shot}  & 58\% & 0.390 &  0.210 &  0.153 &  0.117 &  0.520 &  0.265 &  0.489  & +651\% & +1,282\% & +1,300\%\\[1ex]

\approachcmd{T5}   & 79\% & \textbf{0.547} & 0.369 & 0.295 & 0.243 & \textbf{0.606} & 0.384 & \textbf{0.586}  & +746\% & +1,478\% & +1,820\% \\
\approachcmd{FLAN-T5}  & 82\% & 0.546 & \textbf{0.394} & \textbf{0.319} & \textbf{0.273} & 0.599 & \textbf{0.406} & \textbf{0.586} & +755\% & +1,727\% & +\textbf{2,220}\%\\[1ex]

\hybridapproachcmd{T5}  & - & - & - & - & - & - & - &  - &  +749\% & +1,500\% & +1,860\% \\
\hybridapproachcmd{FLAN-T5}  & - & - & - & - & - & - & - &  - & +\textbf{757}\% & +\textbf{1,746}\% & +\textbf{2,220}\% \\
\bottomrule
    \end{tabular}}
    \caption{{We report reformulation performance (left side of the table) measured in terms of reformulation accuracy, BLEU and ROUGE, and QA performance.
    }}
    \label{tab:model_comparison}
\end{table*}

\paragraph{\textbf{Generation Performance:}}
\Cref{tab:model_comparison} shows the BLEU and ROUGE metric results.
Both \flanxxl and \vicuna, achieve lower performance than \approach  for both zero-shot and few-shot scenarios.
For \flanxxl in the few shot setting the exemplars help to obtain better generation performance with an improvement of 3.4 BLEU-1 points.
While \vicuna underperforms \flanxxl,  few-shot exemplars allow it to significantly improve performance, with an increase of 19 BLEU-1 points. Both LLMs are significantly outperformed by \approach, with an increase of 9 ROUGE-L points.

The \approach results highlight two factors: (1) generative models are better for this task, as the extractive baseline achieves the lowest performance on all metrics. (2) reformulating buyer messages is complex, and accurate performance requires fine-tuning.

\paragraph{\textbf{Reformulation Accuracy:}}
Due to limitations of automated metrics ~\cite{DBLP:conf/acl/YangLLLL18}, we manually assess reformulation accuracy, computed for: \vicunacmd{-fs}, \flancmd{-fs}, \approachcmd{t5} and \approachcmd{ft5}.
\approach obtains the highest reformulation accuracy with 79\% for \approachcmd{T5} and 82\% for \approachcmd{FT5}. 
Note that the size of the base model is important: \approachcmd{FT5} has 3\% higher accuracy than \approachcmd{T5}.
Without fine-tuning, \flanxxl only obtains a 58\% accuracy, a drop of $\blacktriangledown$24\% compared to \approachcmd{FT5}. Similarly, \vicuna obtains a reformulation accuracy of 31\%, a $\blacktriangledown$51\% drop compared to \approachcmd{FT5}. 

\paragraph{\textbf{Question Answering Performance:}}
\Cref{tab:model_comparison} shows the QA results in terms of the understanding confidence scores and answer rates.\footnote{We assess the proportion of questions answered when the QA confidence surpasses a threshold.} For reasons of confidentiality, the QA results are reported as relative improvements over the extractive baseline.
On question understanding, \approachcmd{T5} and \approachcmd{FT5} obtain relative improvements over \heuristicbaseline with 746\% and 755\%, respectively. 
Similarly, on answering rate, \approachcmd{FT5} obtains the highest improvement with 1,727\%, while for \approachcmd{T5} the improvement is 1,478\%. This result shows that the buyer's messages in their original form are unsuitable for QA.

\hybridapproach achieves the highest relative improvement across all metrics. 
This validates our intuition that for shorter messages already in question form, reformulation is not necessary, and in such cases an extractive method provides accurate answers.

Finally, in terms of answer relevance, we see a relative improvement of 2,220\% over the extractive baseline, and see no difference between \approachcmd{FT5} and \hybridapproachcmd{FT5}.
This shows that not only do we increase answer rates, but also answer precision, as the relative increases in relevance from \approach are much higher compared to other baselines.

\section{Online Deployment and Evaluation}

We deployed the more cost-effective \hybridapproachcmd{T5} model, exhibiting a performance equal to \hybridapproachcmd{FT5} according to offline results.

We assess user satisfaction and purchase metrics from millions of e-commerce customers.
We split users into two cohorts: a control group (\control) whose messages are manually answered by sellers, and a treatment group (\treatment),  whose questions are answered by \hybridapproachcmd{T5}.
We also consider \treatmentrel, a subset of \treatment, who provide explicit positive feedback on \approach answers.
We consider the following online evaluation metrics:

\begin{itemize}
    \item \textbf{Purchase Rate -- PR}:
    The ratio of \emph{unique users} who ask a question about a product and buy it within a week, to the total number of users asking a question.

    \item \textbf{Successful Answer Rate -- SAR}: The proportion of messages that received an instant answer where buyers were satisfied and did not send it to the seller.\footnote{If unsatisfied with an instant answer, users can forward their question to the seller.}

\end{itemize}

\subsection{Results}
Due to confidentiality, the results are reported as relative improvements over the control cohort (\control).
Table~\ref{tab:extrinsic_evaluation} shows the results for the PR and SAR rate metrics.

\textbf{Purchase Rate -- PR:} Both treatment groups have significant increases in PR. \treatmentrel obtains the highest purchase rate. This is intuitive given that the instant answers are explicitly marked as helpful.
This result demonstrates that providing instant answers accelerates customer purchase decisions.

\textbf{Successful Answer Rate -- SAR:}
On \treatment cohort, users submit significantly fewer messages to sellers (50\% relative increase of SAR).
For \treatmentrel, SAR increases by 276\%.

\begin{table}[ht!]
    \centering
    \resizebox{0.5\linewidth}{!}{
    \begin{tabular}{l c c c}
         \toprule
 &  Control & \treatment & \treatmentrel\\
\midrule
PR & $0.0$ &  $+28.57$\% & $+50.88$\%\\
SAR & $0.0$ & $+57.14$\% & $+276.73$\% \\
\bottomrule
    \end{tabular}}
    \caption{Online evaluation results with real customers. The reported metrics represent relative improvement over the control cohort ($C$).}
    \label{tab:extrinsic_evaluation}
\end{table}\vspace{-20pt}
\section{Conclusion}\label{sec:conclusion}
We proposed \approach, an approach for automatically answering messages that are sent from buyers to sellers.
Offline experiments validated our approach, and live deployment demonstrated that it improves the shopping experience for both buyers and sellers.

Our method efficiently reformulates messages into concise, salient questions optimal for understanding and response by a federated QA system, providing instant answers to buyers.
The instant answers feature significantly influences both buyers and sellers, evidenced by a reduction of up to 276\% in buyer-to-seller messages. This decrease likely reflects users' satisfaction with the instant responses from \approach, contributing to enhanced buyer experiences and decreased seller overhead. An empirical online study involving real e-commerce users demonstrated a substantial relative increase in purchase rates by 57.14\% when compared to a control group not utilizing \approach instant answers.

\bibliographystyle{splncs04}
\bibliography{references}

\begin{thebibliography}{10}
\providecommand{\url}[1]{\texttt{#1}}
\providecommand{\urlprefix}{URL }
\providecommand{\doi}[1]{https://doi.org/#1}

\bibitem{ahearne2022future}
Ahearne, M., Atefi, Y., Lam, S.K., Pourmasoudi, M.: The future of buyer--seller
  interactions: A conceptual framework and research agenda. Journal of the
  Academy of Marketing Science pp. 1--24 (2022)

\bibitem{cao2022tasa}
Cao, Y., Li, D., Fang, M., Zhou, T., Gao, J., Zhan, Y., Tao, D.: Tasa:
  Deceiving question answering models by twin answer sentences attack. arXiv
  preprint arXiv:2210.15221  (2022)

\bibitem{chen2020jddc}
Chen, M., Liu, R., Shen, L., Yuan, S., Zhou, J., Wu, Y., He, X., Zhou, B.: The
  jddc corpus: A large-scale multi-turn chinese dialogue dataset for e-commerce
  customer service. In: Proceedings of the 12th Language Resources and
  Evaluation Conference. pp. 459--466 (2020)

\bibitem{vicuna2023}
Chiang, W.L., Li, Z., Lin, Z., Sheng, Y., Wu, Z., Zhang, H., Zheng, L., Zhuang,
  S., Zhuang, Y., Gonzalez, J.E., Stoica, I., Xing, E.P.: Vicuna: An
  open-source chatbot impressing gpt-4 with 90\%* chatgpt quality (March 2023),
  \url{https://lmsys.org/blog/2023-03-30-vicuna/}

\bibitem{DBLP:journals/corr/abs-2210-11416}
Chung, H.W., Hou, L., Longpre, S., Zoph, B., Tay, Y., Fedus, W., Li, E., Wang,
  X., Dehghani, M., Brahma, S., Webson, A., Gu, S.S., Dai, Z., Suzgun, M.,
  Chen, X., Chowdhery, A., Narang, S., Mishra, G., Yu, A., Zhao, V.Y., Huang,
  Y., Dai, A.M., Yu, H., Petrov, S., Chi, E.H., Dean, J., Devlin, J., Roberts,
  A., Zhou, D., Le, Q.V., Wei, J.: Scaling instruction-finetuned language
  models. CoRR  \textbf{abs/2210.11416} (2022).
  \doi{10.48550/arXiv.2210.11416},
  \url{https://doi.org/10.48550/arXiv.2210.11416}

\bibitem{cui2017superagent}
Cui, L., Huang, S., Wei, F., Tan, C., Duan, C., Zhou, M.: Superagent: A
  customer service chatbot for e-commerce websites. In: Proceedings of ACL
  2017, system demonstrations. pp. 97--102 (2017)

\bibitem{deng2023product}
Deng, Y., Zhang, W., Yu, Q., Lam, W.: Product question answering in e-commerce:
  A survey. arXiv preprint arXiv:2302.08092  (2023)

\bibitem{do2022cohs}
Do, X.L., Zou, B., Pan, L., Chen, N., Joty, S., Aw, A.: Cohs-cqg: Context and
  history selection for conversational question generation. In: Proceedings of
  the 29th International Conference on Computational Linguistics. pp. 580--591
  (2022)

\bibitem{DBLP:conf/acl/FaustiniCFRM23}
Faustini, P., Chen, Z., Fetahu, B., Rokhlenko, O., Malmasi, S.: Answering
  unanswered questions through semantic reformulations in spoken {QA}. In:
  Sitaram, S., Klebanov, B.B., Williams, J.D. (eds.) Proceedings of the The
  61st Annual Meeting of the Association for Computational Linguistics:
  Industry Track, {ACL} 2023, Toronto, Canada, July 9-14, 2023. pp. 729--743.
  Association for Computational Linguistics (2023),
  \url{https://aclanthology.org/2023.acl-industry.70}

\bibitem{feng2021survey}
Feng, X., Feng, X., Qin, B.: A survey on dialogue summarization: Recent
  advances and new frontiers. arXiv preprint arXiv:2107.03175  (2021)

\bibitem{ferguson2022investigating}
Ferguson, N., Guillou, L., Nuamah, K., Bundy, A.: Investigating the use of
  paraphrase generation for question reformulation in the frank qa system.
  arXiv preprint arXiv:2206.02737  (2022)

\bibitem{gao2021meaningful}
Gao, S., Chen, X., Ren, Z., Zhao, D., Yan, R.: Meaningful answer generation of
  e-commerce question-answering. ACM Transactions on Information Systems (TOIS)
   \textbf{39}(2),  1--26 (2021)

\bibitem{kumar2019question}
Kumar, G., Henderson, M., Chan, S., Nguyen, H., Ngoo, L.: Question-answer
  selection in user to user marketplace conversations. In: 9th International
  Workshop on Spoken Dialogue System Technology. pp. 397--403. Springer (2019)

\bibitem{li2018question}
Li, Y., Miao, Q., Geng, J., Alt, C., Schwarzenberg, R., Hennig, L., Hu, C., Xu,
  F.: Question answering for technical customer support. In: Natural Language
  Processing and Chinese Computing: 7th CCF International Conference, NLPCC
  2018, Hohhot, China, August 26--30, 2018, Proceedings, Part I 7. pp. 3--15.
  Springer (2018)

\bibitem{liao2021practical}
Liao, L.Y., Fares, T.: A practical 2-step approach to assist enterprise
  question-answering live chat. In: Proceedings of the 22nd Annual Meeting of
  the Special Interest Group on Discourse and Dialogue. pp. 457--468 (2021)

\bibitem{lyu2021zero}
Lyu, Q., Zhang, H., Sulem, E., Roth, D.: Zero-shot event extraction via
  transfer learning: Challenges and insights. In: Proceedings of the 59th
  Annual Meeting of the Association for Computational Linguistics and the 11th
  International Joint Conference on Natural Language Processing (Volume 2:
  Short Papers). pp. 322--332 (2021)

\bibitem{mao2020generation}
Mao, Y., He, P., Liu, X., Shen, Y., Gao, J., Han, J., Chen, W.:
  Generation-augmented retrieval for open-domain question answering. arXiv
  preprint arXiv:2009.08553  (2020)

\bibitem{masterov2015canary}
Masterov, D.V., Mayer, U.F., Tadelis, S.: Canary in the e-commerce coal mine:
  Detecting and predicting poor experiences using buyer-to-seller messages. In:
  Proceedings of the Sixteenth ACM Conference on Economics and Computation. pp.
  81--93 (2015)

\bibitem{mcdonald2022detect}
McDonald, T., Tsan, B., Saini, A., Ordonez, J., Gutierrez, L., Nguyen, P.,
  Mason, B., Ng, B.: Detect, retrieve, comprehend: A flexible framework for
  zero-shot document-level question answering. arXiv preprint arXiv:2210.01959
  (2022)

\bibitem{peng2023check}
Peng, B., Galley, M., He, P., Cheng, H., Xie, Y., Hu, Y., Huang, Q., Liden, L.,
  Yu, Z., Chen, W., et~al.: Check your facts and try again: Improving large
  language models with external knowledge and automated feedback. arXiv
  preprint arXiv:2302.12813  (2023)

\bibitem{raffel2020exploring}
Raffel, C., Shazeer, N., Roberts, A., Lee, K., Narang, S., Matena, M., Zhou,
  Y., Li, W., Liu, P.J.: Exploring the limits of transfer learning with a
  unified text-to-text transformer. The Journal of Machine Learning Research
  \textbf{21}(1),  5485--5551 (2020)

\bibitem{rennard2022abstractive}
Rennard, V., Shang, G., Hunter, J., Vazirgiannis, M.: Abstractive meeting
  summarization: A survey. arXiv preprint arXiv:2208.04163  (2022)

\bibitem{samarakoon2011automated}
Samarakoon, L., Kumarawadu, S., Pulasinghe, K.: Automated question answering
  for customer helpdesk applications. In: 2011 6th International Conference on
  Industrial and Information Systems. pp. 328--333. IEEE (2011)

\bibitem{shi2023large}
Shi, F., Chen, X., Misra, K., Scales, N., Dohan, D., Chi, E., Sch{\"a}rli, N.,
  Zhou, D.: Large language models can be easily distracted by irrelevant
  context. arXiv preprint arXiv:2302.00093  (2023)

\bibitem{vakulenko2021question}
Vakulenko, S., Longpre, S., Tu, Z., Anantha, R.: Question rewriting for
  conversational question answering. In: Proceedings of the 14th ACM
  international conference on web search and data mining. pp. 355--363 (2021)

\bibitem{DBLP:conf/acl/YangLLLL18}
Yang, A., Liu, K., Liu, J., Lyu, Y., Li, S.: Adaptations of {ROUGE} and {BLEU}
  to better evaluate machine reading comprehension task. In: Choi, E., Seo, M.,
  Chen, D., Jia, R., Berant, J. (eds.) Proceedings of the Workshop on Machine
  Reading for Question Answering@ACL 2018, Melbourne, Australia, July 19, 2018.
  pp. 98--104. Association for Computational Linguistics (2018).
  \doi{10.18653/v1/W18-2611}, \url{https://aclanthology.org/W18-2611/}

\bibitem{zaib2022conversational}
Zaib, M., Zhang, W.E., Sheng, Q.Z., Mahmood, A., Zhang, Y.: Conversational
  question answering: A survey. Knowledge and Information Systems
  \textbf{64}(12),  3151--3195 (2022)

\end{thebibliography}

\end{document}